\documentclass[10pt]{article}

\usepackage{amsmath,graphicx}
\usepackage{amssymb}
\usepackage{amsmath}
\usepackage{cite}
\usepackage{amsthm}
\usepackage{float}
\usepackage{color,subcaption}
\usepackage{bm}
\usepackage{bbm}
\usepackage{algorithm}
\usepackage{algpseudocode}
\usepackage{enumitem}
\usepackage[margin=1in]{geometry}

\def\defn{\,\triangleq\,}

\def\kbm{{\bm{k}}}

\def\xbm{{\bm{x}}}
\def\ybm{{\bm{y}}}
\def\zbm{{\bm{z}}}
\def\ebm{{\bm{e}}}
\def\rbm{{\bm{r}}}
\def\sbm{{\bm{s}}}
\def\ubm{{\bm{u}}}

\def\Hbm{{\bm{H}}}
\def\Fbm{{\bm{F}}}
\def\Pbm{{\bm{P}}}
\def\Sbm{{\bm{S}}}
\def\Psibm{\bm{\Psi}}
\def\Dbm{{\bm{D}}}

\def\xbmhat{{\widehat{\bm{x}}}}

\def\nablahat{{\hat{\nabla}}}

\def\R{\mathbb{R}}
\def\C{\mathbb{C}}
\def\N{\mathbb{N}}
\def\E{\mathbb{E}}

\def\argmin{\mathop{\mathsf{arg\,min}}}
\def\proposed{\text{PnP-SGD-FPM}}

\def\prox{\mathsf{prox}}
\def\denoise{\mathsf{denoise}}

\newcommand{\diag}[1]{\text{diag}( #1 )}

\theoremstyle{definition}

\begin{document}

\title{Regularized Fourier Ptychography using an \\ Online Plug-and-Play Algorithm}

%%%%%%%%%%%%%%%%%%%%%%%%%%%%%%%%%%%%%%%%%%%%%
%% Authors
%%%%%%%%%%%%%%%%%%%%%%%%%%%%%%%%%%%%%%%%%%%%%

\author{Yu~Sun$^1$,~Shiqi~Xu$^2$,~Yunzhe~Li$^3$,~Lei~Tian$^3$,~Brendt~Wohlberg$^4$,~and~Ulugbek~S.~Kamilov$^{1,2}$\\
\small $^1$\emph{Department of Computer Science and Engineering,~Washington University in St.~Louis, MO 63130, USA.}\\
\small $^2$\emph{Department of Electrical and Systems Engineering,~Washington University in St.~Louis, MO 63130, USA.}\\
\small $^3$\emph{Department of Electrical and Computer Engineering, Boston University, Boston, MA 02215, USA.}\\
\small $^4$\emph{Theoretical Division, Los Alamos National Laboratory, Los Alamos, NM 87545, USA.}\\
}

%\markboth{Deep Learning for Nonlinear Diffractive Imaging}%
%{Sun and Kamilov: Deep Learning for Nonlinear Diffractive Imaging}

\date{}
\maketitle %% required

%%%%%%%%%%%%%%%%%%%%%%%%%%%%%%%%%%%%%%%%%%%%%
%% Abstract
%%%%%%%%%%%%%%%%%%%%%%%%%%%%%%%%%%%%%%%%%%%%%

\begin{abstract}
The plug-and-play priors (PnP) framework has been recently shown to achieve state-of-the-art results in regularized image reconstruction by leveraging a sophisticated denoiser within an iterative algorithm. In this paper, we propose a new online PnP algorithm for Fourier ptychographic microscopy (FPM) based on the fast iterative shrinkage/threshold algorithm (FISTA). 
Specifically, the proposed algorithm uses only a subset of measurements, which makes it scalable to a large set of measurements. We validate the algorithm by showing that it can lead to significant performance gains on both simulated and experimental data. 
\end{abstract}

%%%%%%%%%%%%%%%%%%%%%%%%%%%%%%%%%%%%%%%%%%%%%
%% Introduction
%%%%%%%%%%%%%%%%%%%%%%%%%%%%%%%%%%%%%%%%%%%%%
\section{Introduction}
\label{Sec:Intro}

In computational microscopy, the task of reconstructing an image $\xbm$ of an unknown object from a collection of noisy light-intensity measurements is often formulated as an optimization problem
\begin{equation}
\label{Eq:Optimization}
\xbmhat = \argmin_{\xbm} \left\{f(\xbm)\right\} \quad\text{with}\quad f(\xbm) = d(\xbm) + r(\xbm),
\end{equation}
where the data-fidelity term $d(\cdot)$ ensures the consistency with the measurements and the regularizer $r(\cdot)$ promotes a solution with desirable prior properties such as non-negativity, self-similarity and transform-domain sparsity \cite{Rudin.etal1992, Figueiredo.Nowak2001, Elad.Aharon2006, Danielyan.etal2012}. Due to the nondifferentiability of most regularizers, proximal methods---such as variants of iterative shrinkage/thresholding algorithm (ISTA)~\cite{Figueiredo.Nowak2003, Daubechies.etal2004, Bect.etal2004, Beck.Teboulle2009a} and alternating direction method of multipliers (ADMM)~\cite{Eckstein.Bertsekas1992, Afonso.etal2010, Boyd.etal2011}---are commonly used for solving (\ref{Eq:Optimization}). These algorithms avoid differentiating the regularizer by using a mathematical concept known as the proximal operator, which is itself an optimization problem equivalent to regularized image denoising.

Inspired by this mathematical equivalence, Venkatakrishnan \emph{et al.}~\cite{Venkatakrishnan.etal2013} introduced the plug-and-play priors (PnP) framework for image reconstruction. The key idea in PnP is to replace the proximal operator in an iterative algorithm with a state-of-the-art image denoiser, such as BM3D~\cite{Dabov.etal2007} or TNRD~\cite{Chen.Pock2016}, which does not necessarily have a corresponding regularization objective. Although this implies that PnP methods generally lose interpretability as optimization problems, the framework has gained popularity because of its success in a range of applications in the context of imaging inverse problems~\cite{Sreehari.etal2016, Chan.etal2016, Teodoro.etal2016, Zhang.etal2017a, Teodoro.etal2017, Ono2017, Meinhardt.etal2017, Kamilov.etal2017}. Note that regularization by denoising (RED)~\cite{Romano.etal2017, Metzler.etal2018} is an alternative approach for integrating a denoiser into an imaging problem. The key difference between RED and PnP is that the former builds an explicit regularizer, while the latter relies on an implicit regularization.

%In particular, the effectiveness of PnP was extended beyond the original ADMM formulation~\cite{Venkatakrishnan.etal2013} to other proximal algorithms such as primal-dual splitting and ISTA~\cite{Ono2017, Meinhardt.etal2017, Kamilov.etal2017}. 

Nevertheless, all current iterative PnP algorithms are based on iterative \emph{batch} procedures, which means that they always calculate the updates using the full set of measured data. This makes their application impractical to very large datasets \cite{Bottou.Bousquet2007}. One natural example is Fourier ptychographic microscopy (FPM), where the image formation task relies on hundreds of variably illuminated intensity measurements~\cite{Zheng:2013aa,Tian:15,Zhang:2017aa}.

A novel \emph{online} extension of {PnP-FISTA}, called PnP-SGD, was recently proposed and theoretically analyzed for a convex $d(\cdot)$~\cite{Yu.etal2018}. Here, we extend these results by adapting the algorithm to FPM imaging, where $d(\cdot)$ is nonconvex. We show that the proposed \proposed~enables high-quality FPM imaging at lower computational complexity by using only a small subset of measurements per iteration. We show on both simulated and experimentally measured FPM datasets that the algorithm substantially outperforms its batch counterparts when the memory budget is limited.

\begin{figure*}
\begin{minipage}[t]{.5\textwidth}
\begin{algorithm}[H]
\caption{$\mathsf{PnP}$-$\mathsf{FISTA}$}\label{alg:pnpista}
\begin{algorithmic}[1]
\State \textbf{input: } $\xbm^0 = \sbm^0$, $\gamma > 0$, and $\{q_k\}_{k \in \N}$
\For{$k = 1, 2, \dots$}
\State $\zbm^k \leftarrow \sbm^{k-1}-\gamma \nabla d(\sbm^{k-1})$
\State $\xbm^k \leftarrow \denoise_\sigma(\zbm^k + \sbm^{k-1})$
\State $\sbm^k \leftarrow \xbm^k + ((q_{k-1}-1)/q_k)(\xbm^k-\xbm^{k-1}) $
\EndFor\label{euclidendwhile}
\end{algorithmic}
\end{algorithm}%
\end{minipage}% <---------------- Note the use of "%"
\hspace{0.25em}
\begin{minipage}[t]{.5\textwidth}
\begin{algorithm}[H]
\caption{$\mathsf{PnP}$-$\mathsf{SGD}$}\label{alg:pnpsgd}
\begin{algorithmic}[1]
\State \textbf{input: } $\xbm^0 = \sbm^0$, $\gamma > 0$, $\sigma > 0$, $\{q_k\}$, and $B \geq 1$
\For{$k = 1, 2, \dots$}
%\State $\nablahat d(\sbm^{k-1}) \leftarrow \mathsf{minibatchGradient}(\sbm^{k-1}, B)$
\State $\zbm^k \leftarrow \sbm^{k-1}-\gamma \nablahat d(\sbm^{k-1})$
\State $\xbm^k \leftarrow \denoise_\sigma(\zbm^k)$
\State $\sbm^k \leftarrow \xbm^k + ((q_{k-1}-1)/q_k)(\xbm^k-\xbm^{k-1}) $
\EndFor\label{euclidendwhile}
\end{algorithmic}
\end{algorithm}%
\end{minipage}
\end{figure*}

%%%%%%%%%%%%%%%%%%%%%%%%%%%%%%%%%%%%%%%%%%%%%
%% Background
%%%%%%%%%%%%%%%%%%%%%%%%%%%%%%%%%%%%%%%%%%%%%

\section{Background}
\label{Sec:Background}

\subsection{FPM as an inverse problem}
\label{Sec:InverseProblem}
Consider an unknown object with a complex transmission function $o(\rbm)$, where $\rbm$ denotes the spatial coordinates at the object plane. A total of $N$ LED sources are used to illuminate the object. Each illumination is treated as a local plane wave with a unique spatial frequency $\kbm_i$, $i \in \{1,...,N\}$. The exit wave from the object is described by the product: $u(\rbm) = o(\rbm)e^{i \langle \kbm_i, \rbm \rangle}$, which indicates that the center of the sample's spectrum is shifted to $\kbm_i$~\cite{Zheng:2013aa,Tian:15,Zhang:2017aa}. At the pupil plane, the shifted Fourier transformation of the exit wave is further filtered by the pupil function $p(\kbm)$. For a single illumination, the discrete FPM model can be mathematically described by the following inverse problem
\begin{equation}
\label{Eq:ForwardModel}
\ybm = |\Hbm \sbm(\xbm)|^2 + \ebm,\quad \text{with}\quad \Hbm \defn \Fbm^{-1}_c \diag{\Pbm} \Sbm \Fbm_o
\end{equation}
where $\sbm(\xbm)=e^{i\xbm}$ denotes the discretized transmittance, with $\xbm\in\R^n$ being the vectorized representation of the desired object properties, $\ybm\in\R^m$ represents the corresponding low-resolution light-intensity measurements, and $\ebm$ is the noise vector. The operator $|\cdot|$ computes the element-wise absolute value. The complex matrix $\Hbm \in \C^{m \times n}$ is implemented by taking the Fourier transform ($\Fbm_o$) of the object, shifting and truncating the low frequency region ($\Sbm$), multiplying it by a pupil function in the frequency domain ($\Pbm$), and taking the inverse Fourier transform ($\Fbm^{-1}_c$) with respect to the truncated spectrum.

In practice, the inverse problem is often reformulated as (\ref{Eq:Optimization}) using a quadratic data-fidelity term
\begin{equation}
\label{Eq:InverseProblem}
d(x) = \| \; |\Hbm \ubm(\xbm)|^2 - \ybm\|_2^2.
\end{equation}
Examples of popular regularizers in imaging include the spatial sparsity-promoting penalty $r(\xbm) \defn \lambda \|\xbm\|_{1}$ and total variation (TV) penalty ${r(\xbm) \defn \lambda\|\Dbm\xbm\|_{1}}$, where $\lambda > 0$ is the regularization parameter and $\Dbm$ is the discrete gradient operator~\cite{Rudin.etal1992}. Note that the final objective in FPM is nonconvex due to the absolute value operator in the data-fidelity term~\eqref{Eq:InverseProblem}.

FISTA and ADMM are two widely-used proximal algorithms for dealing with non-differentiable regularizers. The associated key operation is the proximal operator
\begin{equation}
\label{Eq:ProximalOperator}
\prox_{\gamma r}(\zbm) \defn \argmin_{\xbm \in \C^n} \left\{\frac{1}{2}\|\xbm-\zbm\|_2^2 + \gamma r(\xbm)\right\},
\end{equation}
which is mathematically equivalent to an image denoiser formulated as a regularized optimization~\cite{Venkatakrishnan.etal2013}. The algorithms differ from each other in their treatment of the data-fidelity. Whereas FISTA computes the gradient $\nabla d$, ADMM relies on the corresponding proximal operator $\prox_{\gamma d}$. In the context of FPM, we have 
\begin{align}
&\nabla d(\xbm) = [\nabla \Psibm(\xbm)]^\dagger \diag{\Psibm(\xbm)} (\Psibm(\xbm)- \ybm) \\
\text{with} \quad& \Psibm(\xbm) \defn \Hbm \ubm(\xbm),\quad \nabla \Psibm(\xbm) = \diag{\overline{i\text{exp}(i\xbm)}}\Hbm^\dagger,
\end{align}
where $\dagger$ denotes the conjugate transpose, and
\begin{equation}
\label{Eq:ADMMDataUpdate}
\prox_{\gamma d}(\xbm) = \argmin_{\zbm \in \R^n} \left\{\frac{1}{2}\|\zbm-\xbm\|_2^2 + \frac{\gamma}{2}\||\Hbm \ubm(\zbm)|^2-\ybm\|_2^2\right\}.
\end{equation}
Because the data-fit in FPM is not convex, the optimization in~\eqref{Eq:ADMMDataUpdate} is difficult to solve. Additionally, the calculation of (\ref{Eq:ADMMDataUpdate}) is computationally expensive when the number of measurements is large.

\subsection{Denoiser as a prior}

Since the proximal operator is mathematically equivalent to regularized image denoising, the powerful idea of Venkatakrishnan \emph{et al.}~\cite{Venkatakrishnan.etal2013} was to consider replacing it with a more generic denoising operator $\denoise_\sigma(\cdot)$ controlled by ${\sigma > 0}$. In order to be backward compatible with the traditional optimization formulation, this strength parameter is often scaled with the step-size as $\sigma = \sqrt{\gamma \lambda}$, for some parameter $\lambda > 0$.

%Although the original formulation of PnP only considers ADMM\cite{Venkatakrishnan.etal2013}, recent results have shown that it is compatible with other proximal algorithms~\cite{Ono2017, Meinhardt.etal2017, Kamilov.etal2017}. 
Recently, the effectiveness of PnP was extended beyond the original ADMM formulation~\cite{Venkatakrishnan.etal2013} to other proximal algorithms such as primal-dual splitting and ISTA~\cite{Ono2017, Meinhardt.etal2017, Kamilov.etal2017}. The formulation of PnP based on ISTA is summarized in Algorithm~\ref{alg:pnpista}, where we introduce a sequence that controls the shift between FISTA and ISTA. When the following updates is adopted 
%{\scriptsize $q_k \leftarrow \frac{1}{2}(1+\sqrt{1+4q_{k-1}^2})$}
\begin{equation}
q_k \leftarrow \frac{1}{2}(1+\sqrt{1+4q_{k-1}^2}), 
\end{equation}
the FISTA is used. On the other hand, if $q_k$ is set to one for any $k\in\N$, the standard ISTA is considered.  As the FISTA is known to converge faster, we design the \proposed~based on the FISTA formulation of PnP. 

The theoretical convergence of PnP-ISTA was analyzed in a recent paper~\cite{Yu.etal2018} for convex and differentiable $d(\xbm)$. It was shown that the PnP-ISTA converges to a fixed point $\xbm^\ast$ at the rate of $O(1/t)$ if $\denoise_\sigma(\cdot)$ is a $\theta$-averaged operator.

%%%%%%%%%%%%%%%%%%%%%%%%%%%%%%%%%%%%%%%%%%%%%
%% Proposed Method
%%%%%%%%%%%%%%%%%%%%%%%%%%%%%%%%%%%%%%%%%%%%%

\section{Proposed Method}
\label{Sec:Method}

We now introduce the online algorithm \proposed~and describe its advantages over PnP-FISTA. In FPM, the data-fidelity term $d$ consists of a large number of component functions
\begin{equation}
\label{Eq:ComponentData}
d(\xbm) = \E[d_j(\xbm)] = \frac{1}{J}\sum_{j = 1}^J d_j(\xbm),
\end{equation}
where each $d_j$ typically depends only on the subset of the measurements $\ybm$. Note that in the notation~\eqref{Eq:ComponentData}, the expectation is taken over a uniformly distributed random variable ${j \in \{1,\dots, J\}}$. The computation of the gradient of $d$,

%%%%%%%%%%%%%%%%%% Figure 1 %%%%%%%%%%%%%%%%%%%%%%%%
\begin{figure*}[t]
\begin{center}
\includegraphics[width=16cm]{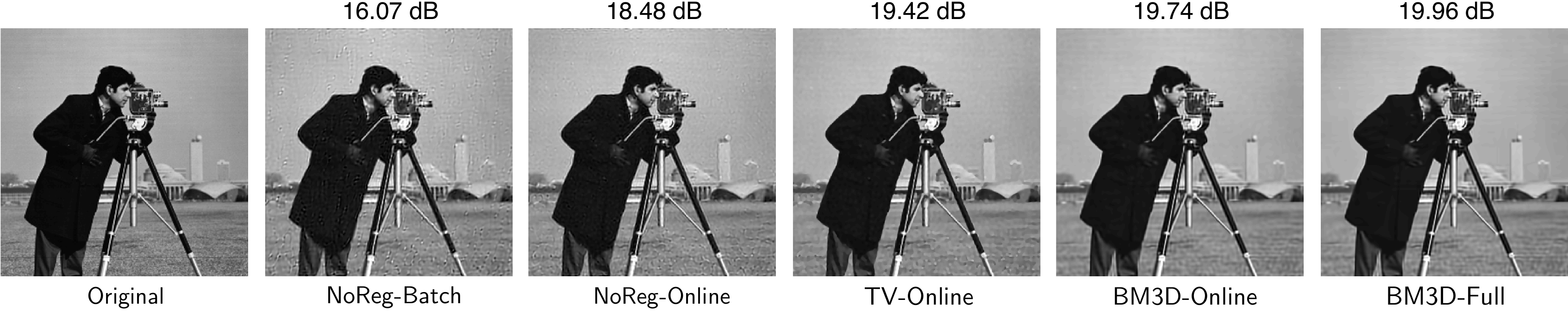}
\end{center}
\caption{Visual comparison of the reconstructed images of \emph{Cameraman} obtained by \proposed~and PnP-FISTA. Minibatch $B=60$ was used in the simulation. The first column ($\mathsf{Original}$) shows the original image. The second column ($\mathsf{NoReg}$-$\mathsf{Batch}$) presents the result of PnP-FISTA using no regularizer. The third ($\mathsf{NoReg}$-$\mathsf{Online}$), fourth ($\mathsf{TV}$-$\mathsf{Online}$) and fifth ($\mathsf{BM3D}$-$\mathsf{Online}$) columns present the results of \proposed~using no regularizer, using TV and using BM3D, respectively. The last column (BM3D-Full) shows the result of PnP-FISTA using BM3D and all 293 measurements. Each image is labeled with its SNR value with respect to the original image.}
\label{Fig:VisualExamples}
\end{figure*}
%%%%%%%%%%%%%%%%%%%%%%%%%%%%%%%%%%%%%%%%%%%%%%%%%%%%%

\begin{equation}
\nabla d(\xbm) = \E[\nabla d_j(\xbm)] = \frac{1}{J} \sum_{j = 1}^J \nabla d_j(\xbm),
\end{equation}
scales with the total number of components $J$, which means that when the latter is large, the classical batch PnP algorithms may become impractical in terms of speed or memory requirements. The central idea of $\proposed$, summarized in Algorithm~\ref{alg:pnpsgd}, is to approximate the gradient at every iteration with an average of $B \ll J$ component gradients
\begin{equation}
\label{Eq:StochGrad}
\nablahat d(\xbm) = \frac{1}{B}\sum_{b = 1}^B \nabla d_{j_b}(\xbm),
\end{equation}
where $j_1, \dots, j_B$ are independent random variables that are distributed uniformly over $\{1, \dots, J\}$. The minibatch size parameter $B \geq 1$ controls the number of gradient components used at every iteration. Note that, when the $d(\xbm)$ is convex and Lipschitz continuous, the \proposed~is shown to converge to a fixed point $\xbm^\ast$ at the worst-case rate of $O(1/\sqrt{t})$~\cite{Yu.etal2018}.

%%%%%%%%%%%%%%%%%% Figure 2 %%%%%%%%%%%%%%%%%%%%%%%%
\begin{figure}[t]
\begin{center}
\includegraphics[width=8.5cm]{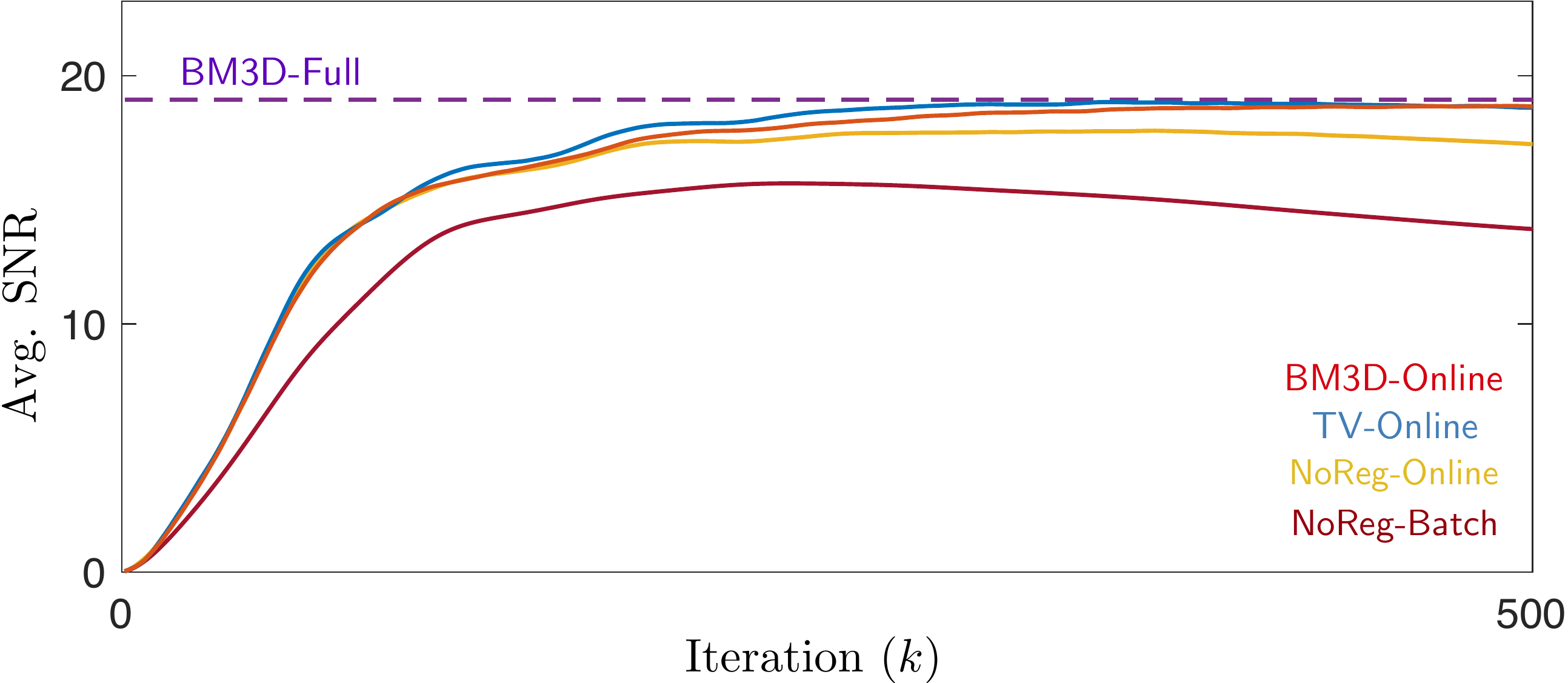}
\end{center}
\caption{Evolution of average SNR across iterations for batch and online PnP algorithms using different priors. The corresponding labels are shown at the bottom-right corner inside the plot. The purple dotted line, BM3D-Full, indicates the performance when using all the available measurements. Note that BM3D-Online achieves the SNR performance of BM3D-Full at a lower computational cost.}
\label{Fig:SnrCompare}
\end{figure}
%%%%%%%%%%%%%%%%%%%%%%%%%%%%%%%%%%%%%%%%%%%%%%%%%%%%%

%%%%%%%%%%%%%%%%%%%%%%%%%%%%%%%%%%%%%%%%%%%%%
%% Experiments
%%%%%%%%%%%%%%%%%%%%%%%%%%%%%%%%%%%%%%%%%%%%%

\section{Numerical Validation}
\label{Sec:Validation}
In this section, we validate our \proposed~on simulated and experimental data by considering two representative denoisers: TV~\cite{Rudin.etal1992} and BM3D~\cite{Danielyan.etal2012}. Note that our focus is to demonstrate the effectiveness of the proposed PnP method for FPM rather than to test different denoisers, although the algorithm is readily compatible with other state-of-the-art denoising methods.

\subsection{Experimental Setup}
\label{Sub:FpmSystem}
We set up the simulation to match the FPM system used for the experimental data~\cite{Tian:15}. The sample is placed 70 mm below a 32 $\times$ 32 surface-mounted LED array with a 4 mm pitch. All LED sources generate light with a wavelength 513 mm and the bandwidth 20 nm. We individually illuminate the sample with 293 LEDs centered in the array and record the corresponding intensity measurements with a camera placed under the sample. The numerical aperture (NA) of the objective is 0.2~\cite{Tian:15}. We achieve the synthetic NA of around 0.7 by summing the NA of the objective and illumination. 
%We use eight common gray-scale images with the size of $500 \times 500$ pixels as samples in the simulations, while the open dataset of HeLa cells provided in~\cite{Tian:15} are considered in experimental evaluations. 

%%%%%%%%%%%%%%%%%% Figure 1 %%%%%%%%%%%%%%%%%%%%%%%%
\begin{figure}[t]
\begin{center}
\includegraphics[width=8.6cm]{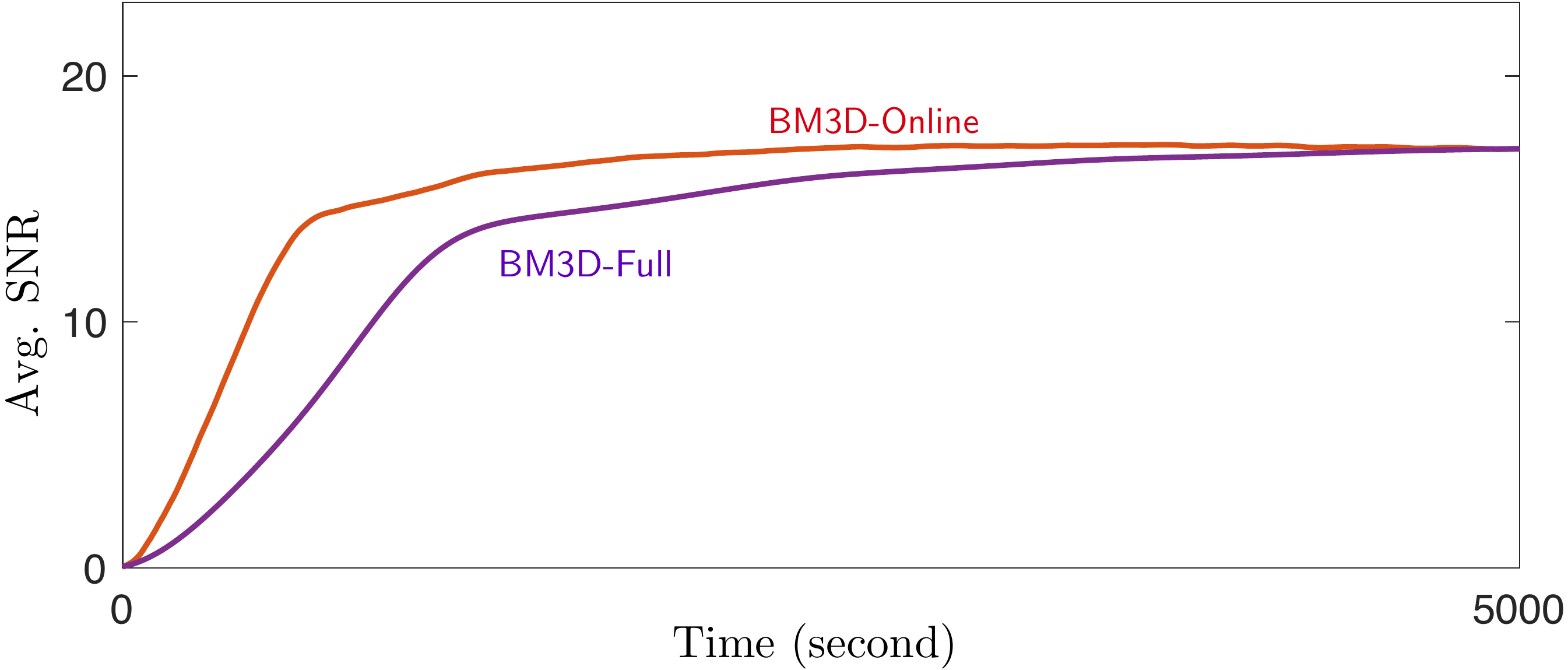}
\end{center}
\caption{Comparison between BM3D-Online and BM3D-Full for a fixed reconstruction time. The average SNR is plotted against the time in seconds for both algorithms. BM3D-Online uses only 60 measurements per iteration, while BM3D-Full uses all 293 measurements. The lower per iteration cost leads to a substantially faster convergence of BM3D-Online.}
\label{Fig:TimeCompare}
\end{figure}
%%%%%%%%%%%%%%%%%%%%%%%%%%%%%%%%%%%%%%%%%%%%%%%%%%%%%

\subsection{Benefits of \proposed}
We first quantitatively analyze the performance of \proposed~by reconstructing six common gray-scale images discretized to ${500 \times 500}$ pixels: \emph{Cameraman}, \emph{House}, \emph{Jet}, \emph{Lenna}, \emph{Pepper} and \emph{Woman}. The simulated measurements were obtained by solving the forward model defined in (\ref{Eq:ForwardModel}). Additionally, the measurements were corrupted by an additive Gaussian noise (AWGN) corresponding to 40 dB of input signal-to-noise ratio (SNR). The SNR is also used as a metric for numerically evaluating the quality of reconstruction. In simulations, all algorithmic hyperparameters were optimized for the best SNR performance with respect to the original image. We use \emph{average SNR} to indicate the SNR averaged over all the test images.

%%%%%%%%%%%%%%%%%% Figure 1 %%%%%%%%%%%%%%%%%%%%%%%%
\begin{figure*}[t]
\begin{center}
\includegraphics[width=16cm]{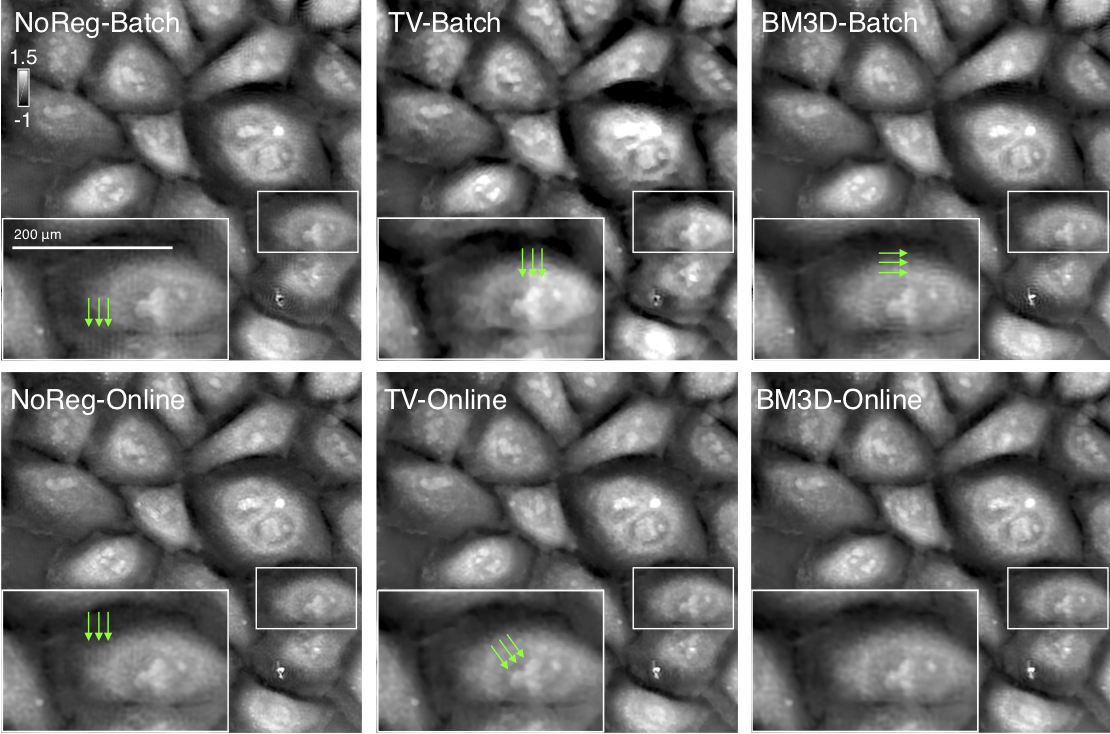}
\end{center}
\caption{Comparison of online and batch algorithms on the FPM dataset containing HeLa cells. Each algorithm uses 60 measurements per-iteration. The first row illustrates the results of $\mathsf{NoReg}$-$\mathsf{Batch}$, $\mathsf{TV}$-$\mathsf{Batch}$, and $\mathsf{BM3D}$-$\mathsf{Batch}$. The second shows $\mathsf{NoReg}$-$\mathsf{Online}$, $\mathsf{TV}$-$\mathsf{Online}$, and $\mathsf{BM3D}$-$\mathsf{Online}$. Visual difference are illustrated by the white rectangles drawn inside the images. The green arrows highlight the artifacts.}
\label{Fig:HeLa}
\end{figure*}
%%%%%%%%%%%%%%%%%%%%%%%%%%%%%%%%%%%%%%%%%%%%%%%%%%%%%

Figure~\ref{Fig:SnrCompare} illustrates the evolution of average SNR across iterations for different priors and PnP variants. With the minibatch $B=60$, the online algorithms randomly select different subset of measurements at each iteration, whereas the batch algorithms use the same measurements in the reconstruction. Hence, as expected, by eventually cycling through all the measurements, NoReg-Online achieves a higher average SNR than NoReg-Batch, which only uses the fixed $60$ illuminations. Additionally, the performance is further enhanced by using priors. For example, TV-Online and BM3D-Online increase the average SNR from 18.48 dB to 19.42 dB and 19.74 dB, respectively. A visual illustration on \emph{Cameraman} is presented in Figure~\ref{Fig:VisualExamples}, where BM3D-Full is shown as reference. We observe that, even with $B=60$, the solution of BM3D-Online is only 0.25 dB lower than the batch PnP algorithm using all 293 measurements.

Figure~\ref{Fig:TimeCompare} compares the average SNR performance of online and batch PnP algorithms within a fixed run-time. In the test, BM3D-Online uses only 60 measurements per iteration, while BM3D-Full uses all 293 measurements. The lower per iteration cost makes the reconstruction of BM3D-Online substantially faster than that of its batch rival. In particular, the averaged single-iteration run-time of BM3D-Online and BM3D-Full was 9.07 seconds and 19.66 seconds, respectively. We also note that BM3D-Online and BM3D-Full eventually converge to the same average SNR, which agrees with the plots in Figure~\ref{Fig:SnrCompare}. Additionally, \proposed~achieves a substantial speedup due to its reduction in per-iteration computational complexity, which makes the algorithm applicable to very large datasets.

\subsection{Validation on real data}
We now validate the performance of \proposed~on experimental FPM data. The sample used in the experiment consists of the human cervical adenocarcinoma epithelial (HeLa) cells~\cite{Tian:15}. The system corresponds to the FPM described in Section \ref{Sub:FpmSystem} with total 293 measurements for reconstruction. 

Figure~\ref{Fig:HeLa} compares the images of HeLa cells reconstructed by \proposed~and PnP-FISTA. Each image has the resolution of $900 \times 900$ pixels. The green arrows in the white rectangles highlight the artifacts. We consider the scenario with a limited memory budget sufficient only for 60 measurements. Since batch algorithms use a fixed subset of measurements, they produce strong unnatural features, such as the horizontal streaking artifacts in NoReg-Batch. Both TV-Batch and BM3D-Batch mitigate this artifact by using priors, but they generate blockiness and vertical grids in the cell, respectively. Online algorithms generally improve the visual quality by making use of all the data. Finally, BM3D-Online alleviates the artifacts and reconstructs a high-quality image, while NoReg-Online and TV-Online still have streaking artifacts and blockiness.

\section{Conclusion}
\label{Sec:Conclution}
In this paper, we propose a novel online plug-and-play algorithm for the regularized Fourier ptychographic microscopy. Numerical simulations demonstrate that \proposed~converges to a nearly-optimal average SNR in a shorter amount of time. The experiments for FPM confirm the effectiveness and efficiency of \proposed~in practice, and shows its potential for other imaging applications. More generally, this work shows the potential of \proposed~to solve inverse problems beyond traditional convex data-fidelity terms.

%%%%%%%%%%%%%%%%%%%%%%%%%%%%%%%%%%%%%%%%%%%%%
%% References
%%%%%%%%%%%%%%%%%%%%%%%%%%%%%%%%%%%%%%%%%%%%%

\end{document}